# Application of Classification and Feature Selection in Building Energy Simulations

Fatemeh Shahsavari[1], Zohreh Shaghaghian[1]

[1] Texas A&M,
College Station, TX, United States
fatemeh.shahsavari@tamu.edu

**ABSTRACT**

Building energy performance is one of the key features in performance-based building design decision making. Building envelope materials can play a key role in improving the building energy performance. The thermal properties of building materials determine the level of heat transfer through building envelope, thus the annual thermal energy performance of the building. This research applies the Linear Discriminant Analysis (LDA) method to study the effects of materials' thermal properties on building thermal loads. Two approaches are adopted for feature selection including the Principal Component Analysis (PCA) and the Exhaustive Feature Selection (EFS). A hypothetical design scenario is developed with six material alternatives for an office building in Los Angeles, California. The best design alternative is selected based on the LDA results and the key input parameters are determined based on the PCA and EFS methods. The PCA results confirm that among all thermal properties of the materials, the four parameters including thermal conductivity, density, specific heat capacity, and thickness are the most critical features, in terms of building thermal behavior and thermal energy consumption. This result matches quite well with the assumptions of most of the building energy simulation tools.

## 1   INTRODUCTION

According to [1], the building industry consumes about 40% of total energy in developed countries including the United States. The fossil fuels used for energy in the buildings is the primary source of carbon emissions in the world [2]. These facts necessitate reconsidering building design and construction strategies to improve building energy performance and reduce the environmental effects of the building industry.

In this paper, a simple machine learning classification method known as the Linear Discriminant Analysis (LDA) is used to classify buildings' energy consumption into three levels of low, medium and high. LDA is a simple linear classification method that despite Logistic regression is able to classify more than two categories. Two feature selection methods including the Principal Component Analysis (PCA) and the Exhaustive Feature Selection (EFS) are applied to reduce data dimensionality and find the key features to improve building energy performance. PCA is a feature selection method with a set of vectors, each corresponding to a feature while being orthogonal to the other features. This method produces linear combinations of the features to generate the axes (known as principal components or PCs) [3]. EFS is a wrapper method based on the exhaustive search. In this method, all combinations of features are tested to find the best set [4].

## 2   BACKGROUND

Architectural design decision-making begins with identifying the design problems and objectives. Design decision makers set boundaries for the potential problem solving methods. Several deterministic and non-deterministic building design decision-making approaches such as Simple Multi-attribute Rating Technique (SMART), Analytical Hierarchy Process (AHP) and Analytical Network Process (ANP) are discussed by [5]. There are existing works that have studied the integration of the AHP approach with uncertainty and sensitivity analyses techniques to improve the process of building design decision making [6]. Many studies have investigated data driven approaches using machine learning and advanced deep learning models to explore the advantages of such methods in predicting buildings energy/lighting demand [7], [8]. This study specifically focuses on the building envelope materials to identify the important features that have critical impact on building energy performance.

## 3   DATA PROCESSING AND METHODS

A hypothetical office building located in Los Angeles, CA is used to generate the input dataset to conduct the LDA classification and the feature selection experiment. A modular building design, known as Sprout Space, introduced by [9] (shown in Figure 1) is used in this research.

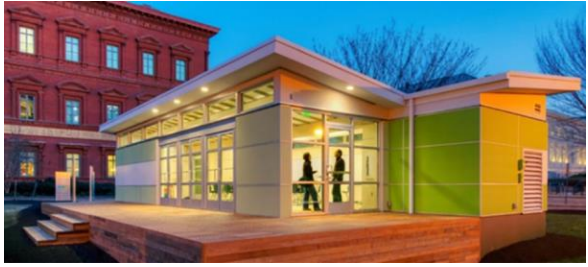

**Figure 1.** Modular building design, Sprout Space [9].

A parametric model of the building is developed in an Architectural model tool, called as Rhinoceros using its visual programming environment, called Grasshopper (shown in Figure 2).

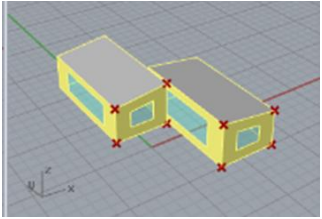

**Figure 2.** Parametric model of the Sprout Space.

In this study, six building materials including timber insulated panel-OSB, timber insulated panel-insulation, concrete, brick, aluminum, and glass are selected to develop the design alternatives. The energy simulation outputs are recorded and used to classify each building design alternative as a high energy design (HED), medium energy design (MED), or low energy design (LED) alternative.

The dataset contains seven features corresponding to the building envelope materials properties of 600 building design alternatives. The seven material properties include thickness, density, thermal conductivity, specific heat capacity, solar absorptance, visual absorptance, and thermal absorptance. The mean and standard deviation for each feature of the building materials are obtained from the literature including [5], [10], [11]. The mean and standard deviation of each feature are used to generate 100 samples from Normal distributions, N ($\mu$, $\sigma^2$) for each material using Monte Carlo sampling method. To choose the sample size, multiple experiments were conducted, and we observed that the samples have a periodic behavior with 100 intervals, thus it showed that 100 samples of each feature would be representing the sampling behavior quite well. Table 1 shows the mean and standard deviation of the input features of the six selected materials.

|  | Timber Insulated Panel - OSB | | Timber Insulated Panel - Insulation | | Concrete | | Brick | | Aluminum | | Glass | |
| --- | --- | --- | --- | --- | --- | --- | --- | --- | --- | --- | --- | --- |
|  | $\mu$ | $\sigma$ | $\mu$ | $\sigma$ | $\mu$ | $\sigma$ | $\mu$ | $\sigma$ | $\mu$ | $\sigma$ | $\mu$ | $\sigma$ |
| **Thickness(m)** | 0.01 | 0.001 | 0.09 | 0.009 | 0.21 | 0.021 | 0.16 | 0.016 | 0.14 | 0.014 | 0.31 | 0.031 |
| **Density (kg/m³)** | 545 | 20 | 11 | 1 | 2000 | 30 | 1700 | 297.5 | 6278 | 2876 | 2509 | 105 |
| **Thermal Conductivity (W/mK)** | 0.135 | 0.0075 | 0.0465 | 0.005 | 1.13 | 0.1 | 0.84 | 0.27 | 244 | 107 | 1.294 | 0.69 |
| **Specific Heat Capacity (J/kg·°K)** | 1740 | 442.5 | 805 | 17.5 | 1000 | 106 | 800 | 86 | 544 | 233 | 820 | 50 |
| **solar absorptance** | 0.5 | 0.05 | 0.5 | 0.05 | 0.5 | 0.05 | 0.5 | 0.05 | 0.5 | 0.05 | 0.5 | 0.05 |
| **visual absorptance** | 0.5 | 0.05 | 0.5 | 0.05 | 0.5 | 0.05 | 0.5 | 0.05 | 0.5 | 0.05 | 0.5 | 0.05 |
| **thermal absorptance** | 0.5 | 0.05 | 0.5 | 0.05 | 0.5 | 0.05 | 0.5 | 0.05 | 0.5 | 0.05 | 0.5 | 0.05 |

**Table 1.** The probability distribution of thermal properties of the selected building materials.

For each design option, the external walls are assigned to one of the materials mentioned above (Table 1). The properties of building materials are the only design variables with uncertainties in this study. Other design variables including glazing U-value, internal heat gains, ventilation and infiltration rates, and building operation schedules are assumed to be fixed. This assumption helps simplify the experiment and keep the focus on testing and demonstrating the applications of classification and feature selection in building energy simulations.

The system variables including the internal heat gain sources, the infiltration rate, and the ventilation rates are listed in Table 2. These values are obtained from the literature [9].

| Variable [Unit] | μ |
|---|---|
| Equipment load per area [W/m$^2$] | 10.98 |
| Infiltration (air flow) rate per area [m$^3$/s-m$^2$] | 0.0003 |
| Lighting density per area [W/m$^2$] | 9.36 |
| Number of people per area [ppl/m$^2$] | 0.25 |
| Ventilation per area [m$^3$/s-m$^2$] | 0.0006 |
| Ventilation per person [m$^3$/s-m$^2$] | 0.005 |

**Table 2.** Description of the system's design assumptions.

The glazing U-value is assumed to be 0.6 W/m$^2$K. The occupancy, lighting, and equipment schedules are matched with the school schedules in the ASHRAE 90.1-2010 [12].

Due to a small size of the building, a single thermal zone is defined for each rectangular section of the building. An ideal loads air system is used for all the thermal zones in the building. The ideal loads air system is an input object which provides a model for an ideal HVAC system. The object is modeled as an ideal VAV terminal unit with variable supply temperature and humidity. The supply air flow rate is varied between zero and the maximum in order to satisfy the zone heating or cooling load, zone humidity controls, outdoor air requirements, and other constraints, if specified [13].

The input samples for each building material are generated to run 600 energy simulations in total. The EnergyPlus energy simulation tool available in Ladybug tools for Grasshopper is used to run the energy simulations for the generated samples. Figure 3 shows the distribution of the 600 samples of the data showing the building thermal loads [kWh/m$^2$] (the sum of annual cooling load [kWh/m$^2$] and heating load [kWh/m$^2$]) resulting from each type of the envelope material.

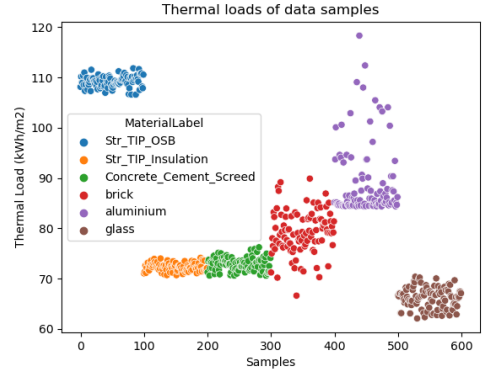

**Figure 3.** The building annual thermal load results corresponding to a batch of 100 samples for each envelope material.

For further analysis, the data are split to the training and test samples with a ratio of 35% and 65 %, respectively. The small percentage of the training data compared to the test data reduces the potential risk of overfitting. Also, the data are normalized due to having different scales of the features. The LDA classification method and two methods of feature selection including the Principal Component Analysis (PCA) and the Exhaustive feature selection (EFS) are implemented. The results are discussed in the next section.

## 4 RESULTS AND DISCUSSION

The LDA classification method combined with the PCA feature selection is conducted on the training data (210 samples). The scree plots are shown in Figure 4 and Figure 5 to show the percentage of variance corresponding to each principal component (PC). The PCs are a sequence of vectors, where each vector is the direction of a line that best fits the data while being orthogonal to the other vectors [3].

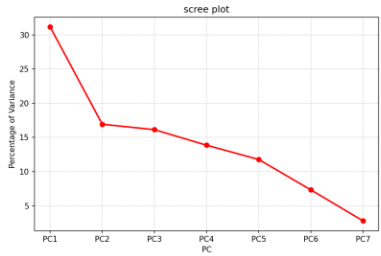

**Figure 4.** The scree plot: The percentage of variance for PC1 to PC7.

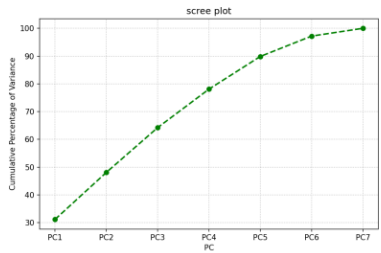

**Figure 5.** Cumulative Percentage of Variance for PC1 to PC7.

The scree plots show that PC1 can explain 30% of the data variance. The cumulative plot reveals that at least five PCs are needed to determine 95% of the variance of the data. The plots also show that PC1 can perform a fair job in discriminating samples with low, medium, and high thermal loads.

The results are plotted in Figure 6, Figure 7, and Figure 8 showing the pairs of the first three Principal Components (PCs).

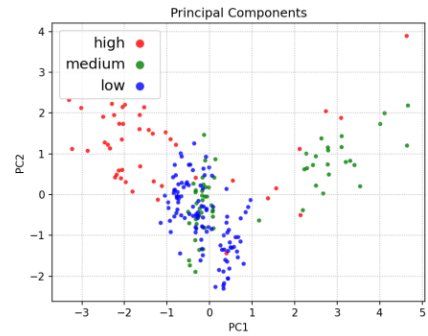

**Figure 6.** PC1 vs. PC2.

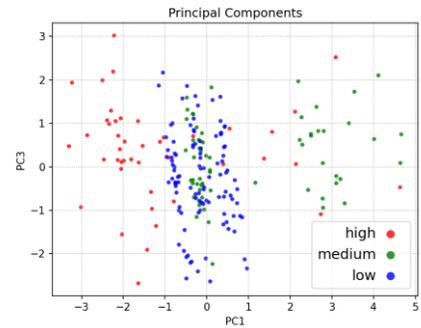

**Figure 7.** PC1 vs. PC3.

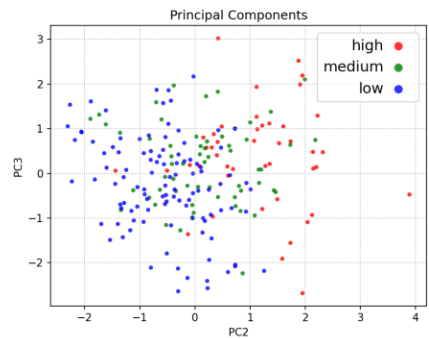

**Figure 8.** PC2 vs. PC3.

In the plots shown in Figure 6, Figure 7, and Figure 8, the samples are divided into three groups based on their corresponding simulation outputs. The simulation output points are categorized with "high" labels if the energy consumption result is equal to or more than 90 kWh/m$^2$, and "low" labels if the energy result is equal or less than 75 kWh/m$^2$. The data is labeled as "medium" if the energy result is higher than 75 kWh/m$^2$ and less than 90 kWh/m$^2$. Note that 75 kWh/m$^2$ and 90 kWh/m$^2$ are only the thresholds specified for the purpose of the classification in this study. Since the input assumptions for this experiment have been simplified to keep the focus of the paper on the applications of machine learning methods, thus the relatively high simulation outputs may be a result of this simplification in the input variable assumptions.

The feature selection using the PCA method resulted in the loading matrix shown in Table 3. Note that the absolute values for each feature is shown here.

| Column1 | PC1 | PC2 | PC3 | PC4 | PC5 | PC6 | PC7 |
|---|---|---|---|---|---|---|---|
| Thickness | 0.42 | 0.71 | 0.12 | 0.47 | 0.05 | 0.25 | 0.16 |
| Conductivity | 0.80 | 0.39 | 0.15 | 0.31 | 0.01 | 0.01 | 0.30 |
| Specific Heat Capacity | 0.76 | 0.33 | 0.05 | 0.02 | 0.06 | 0.55 | 0.11 |
| Density | 0.86 | 0.15 | 0.20 | 0.04 | 0.04 | 0.37 | 0.26 |
| Thermal absorptance | 0.09 | 0.27 | 0.65 | 0.52 | 0.48 | 0.09 | 0.01 |
| Solar absorptance | 0.18 | 0.51 | 0.31 | 0.62 | 0.48 | 0.09 | 0.01 |
| Visual absorptance | 0.13 | 0.28 | 0.73 | 0.05 | 0.61 | 0.05 | 0.02 |

**Table 3.** The loading matrix.

Based on the absolute values shown in Table 3, the following four features have comparatively large coefficients in this analysis:

1. Density (absolute value: 0.86),
2. Thermal Conductivity (absolute value: 0.80),
3. Specific Heat Capacity (absolute value: 0.76), and
4. Thickness (absolute value: 0.42).

Following the results of the feature selection, the classifications of training data using six pairs of the selected features ((a) Specific heat capacity vs. conductivity, (b) Density vs. conductivity, (c) Thickness vs. conductivity, (d) Thickness vs. density, (e) Specific heat capacity vs. density, and (f) Specific heat capacity vs. thickness) are demonstrated in Figure 9-Figure 14.

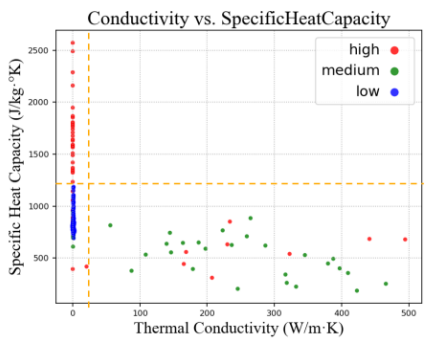

**Figure 9.** The classifications of the training data based on the possible pairs of specific heat capacity and conductivity.

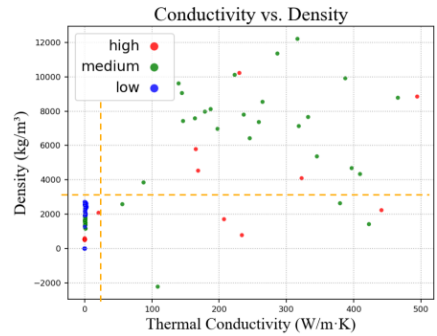

**Figure 10.** The classifications of the training data based on the possible pairs of density and conductivity.

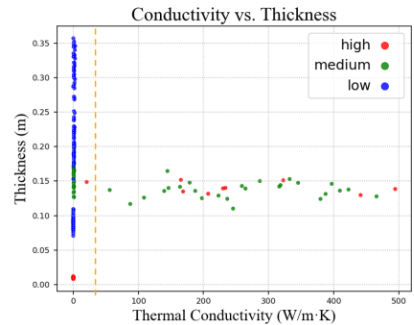

**Figure 11.** The classifications of the training data based on the possible pairs of thickness and conductivity.

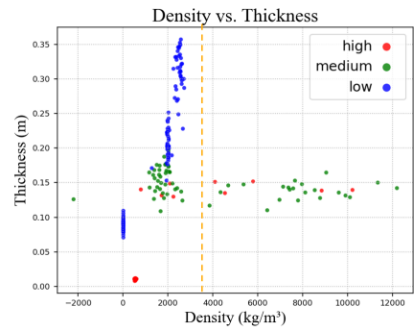

**Figure 12.** The classifications of the training data based on the possible pairs of thickness and density.

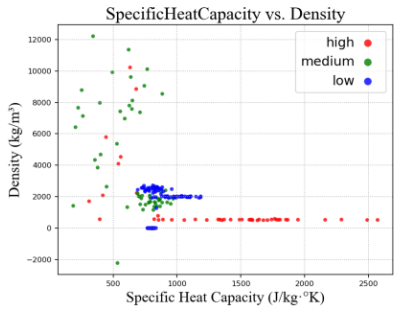

**Figure 13.** The classifications of the training data based on the possible pairs of density and specific heat capacity.

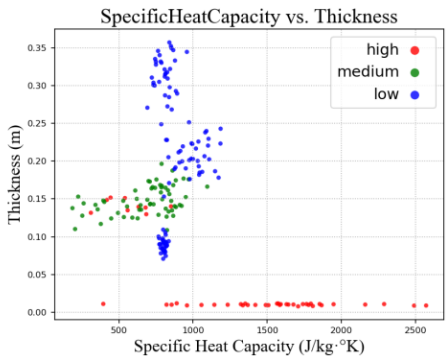

**Figure 14.** The classifications of the training data based on the possible pairs of thickness and specific heat capacity.

Figure 9 shows that in the combination of the specific heat capacity and the thermal conductivity, the former parameter plays a key role in determining building thermal loads. A low level of specific heat capacity results in low energy results, but high energy results are obtained with higher levels of specific heat capacity. Increasing the thermal conductivity results mostly in medium energy results, and high energy results in a few cases. Figure 10 shows that with combination of the density and the thermal conductivity, lower conductivity results in low energy results, while higher conductivity yields medium and high energy results. Figure 11 shows that in the combination of thickness and thermal conductivity, lower conductivity results in low energy results, while higher conductivity yields medium and high energy results. Figure 12 shows that with a combination of the thickness and the density, lower density results in low energy results, while higher conductivity yields medium and high energy results. High thickness results in low energy results. Figure 13 shows that in the combination of specific heat capacity and density, lower density results in high energy results, while higher density yields mostly low and medium energy results. Figure 14 shows that in the combination of specific heat capacity and thickness, lower thickness results in high energy results, while higher thickness yields mostly low and medium energy results.

The results obtained from the PCA method matches to the authors' expectation, since the four features selected by the PCA method are the same variables as most of the energy software including EnergyPlus require in order to run building energy simulations.

The second method of feature selection applied in this study is the Exhaustive Feature Selection (EFS).

Figure 15 shows the accuracy of the LDA classifier conducted on the training data set based on the different combinations of the features using the Exhaustive search.

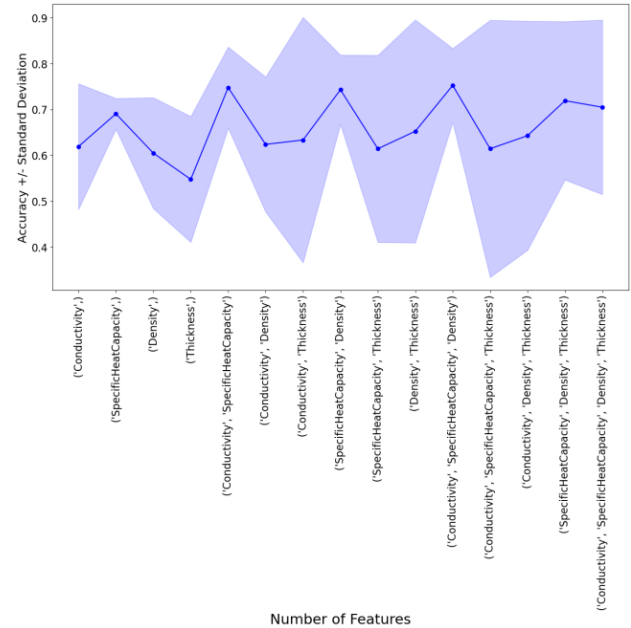

**Figure 15.** The accuracy of the LDA classifier for different combinations of the selected features.

As Figure 15 shows, if selecting only one feature, the key input parameter is the specific heat capacity. The best combination of two features includes the conductivity and the specific heat capacity. The best combination of three features consists of materials' conductivity, specific heat capacity, and density. the LDA classifier for the combination of four features is maximum when choosing conductivity, specific heat capacity, density, and Thermal absorptance.

To validate the findings from the two feature selection methods, the test data are used. First, a classifier trained on the four features selected by the EFS is tested, and then a classifier trained on the four features selected by the PCA method.

The model accuracy on the training data selecting the PCA method is 0.78, while this number is 0.76 when using the EFS method. The model accuracy on the test data is 0.77, when using the PCA method, while it reduces to 0.71 with the EFS method.

The test result confirms that the four selected features by the PCA method result in a better model accuracy on the training and the test data. The relatively low accuracy on the training data means that there is an underfitting that could be explained by the small number of the training samples (only 210). However, the results on the test data show that the model is not overfitted and the accuracy on the test is acceptable. The model accuracy on the training and test data may improve using other classification methods, also with larger number of samples.

# 4 Conclusion

This paper investigates the effects of building envelope materials' thermal properties on the building annual thermal loads, using machine learning classification and feature selection methods. A hypothetical small office building in Los Angeles, CA is selected as a test case. Six different building materials are selected, and seven characteristics of each material are compared in terms of their effects on the building thermal loads. 35% of the data was separated as the training data and the rest of the data was used as the test data. Two methods of Principal Component Analysis (PCA) and Exhaustive Feature Search (EFS) combined with the Linear Discriminant Analysis (LDA) classification method are compared.

The EFS result mostly matches with the PCA result, except for the last feature, that is found to be the thermal absorptance instead of the thickness. The best four features selected by the EFS method are conductivity, specific heat capacity, density, and thermal absorptance. Therefore, upon our prior knowledge about the key features of building materials used in building energy performance simulations, it is recommended to choose PCA method over the EFS method for feature selection in this experiment. This contradiction could be due to the small number of the training data.

This paper concludes that the mentioned machine learning methods could be useful in feature selection and classification for building energy simulations. Further research is recommended to study other machine learning classification and feature selection methods in the field of building energy analysis to distinguish the differences and benefits of each method. Furthermore, since other design input variables including the climate and mechanical systems are know as key parameters affecting the building performance, further research is recommended to study the applications of machine learning methods in different climatic zones, considering different HVAC systems.